\documentclass{article}
\usepackage{ijcai18}

\usepackage{times}
\usepackage{xcolor}
\usepackage{soul}
\usepackage[utf8]{inputenc}
\usepackage[small]{caption}

\usepackage{array,multirow,graphicx}
\usepackage{epsfig}
\usepackage{amsmath}
\usepackage{amsfonts}
\usepackage[linesnumbered,algoruled,boxed,lined]{algorithm2e}

\title{Building Sparse Deep Feedforward Networks using Tree Receptive Fields}

\author{
Xiaopeng Li\thanks{Equal contribution.}, 
Zhourong Chen\footnotemark[1],
Nevin L. Zhang
\\ 
Department of Computer Science and Engineering\\
The Hong Kong University of Science and Technology\\
Clear Water Bay, Hong Kong \\
\{xlibo, zchenbb, lzhang\}@cse.ust.hk
}

\begin{document}

\maketitle

\begin{abstract}
  Sparse connectivity is an important factor behind the success of convolutional neural networks and recurrent neural networks. In this paper, we consider the problem of learning sparse connectivity for feedforward neural networks (FNNs). The key idea is that a unit should be connected to a small number of units at the next level below that are strongly correlated. We use Chow-Liu's algorithm to learn a tree-structured probabilistic model for the units at the current level, use the tree to identify subsets of units that are strongly correlated, and introduce a new unit with receptive field over the subsets. The procedure is repeated on the new units to build multiple layers of hidden units. The resulting model is called a TRF-net. Empirical results show that, when compared to dense FNNs, TRF-net achieves better or comparable classification performance with much fewer parameters and sparser structures. They are also more interpretable.
\end{abstract}

\section{Introduction}
It is common knowledge that architecture matters in deep learning. In general, convolutional neural networks (CNNs) are considered suitable for spatial data and recurrent neural networks (RNNs) are considered suitable for sequential data. However, it remains an art to design an appropriate architecture for a particular application. Typically, researchers need to evaluate a long list of candidate architectures before finding a satisfactory one, a process that is sometimes called graduate student descent. For this reason, there is growing interest in architecture learning \cite{baker2016designing,zoph2016neural}.

In addition to saving manual labor, architecture learning is interesting for two other reasons. First, it enables us to detect salient patterns in data and represent them using network structures. This can lead to better performing and more interpretable models. Second, architecture learning naturally would involve connectivity learning, which leads to sparse models. Model sparsity is helpful in avoiding overfitting and is desirable for applications on hand-held devices. 

In this paper, we focus on feedforward neural networks (FNNs). While CNNs are designed for spatial data and RNNs for sequential data, FNNs are used for vector-form data that are neither spatial nor sequential. By definition, FNNs consists of fully-connected layers, where a neuron is connected to all neurons at the next level below. We call such models dense FNNs.  Our objective is to learn sparse FNNs where a neuron is connected to only a small number of neurons at the next level below. One way to achieve this is to first learn a fully-connected model and then prune weak links \cite{han2015learning,guo2016dynamic,Srinivas2015,li2016pruning}. This approach does not solve the architecture learning problem because one needs to determine the structure for the fully-connected model to begin with. We aim at learning sparse FNN structures from scratch.

\begin{figure}[t]
\begin{center}
\includegraphics[width=\linewidth]{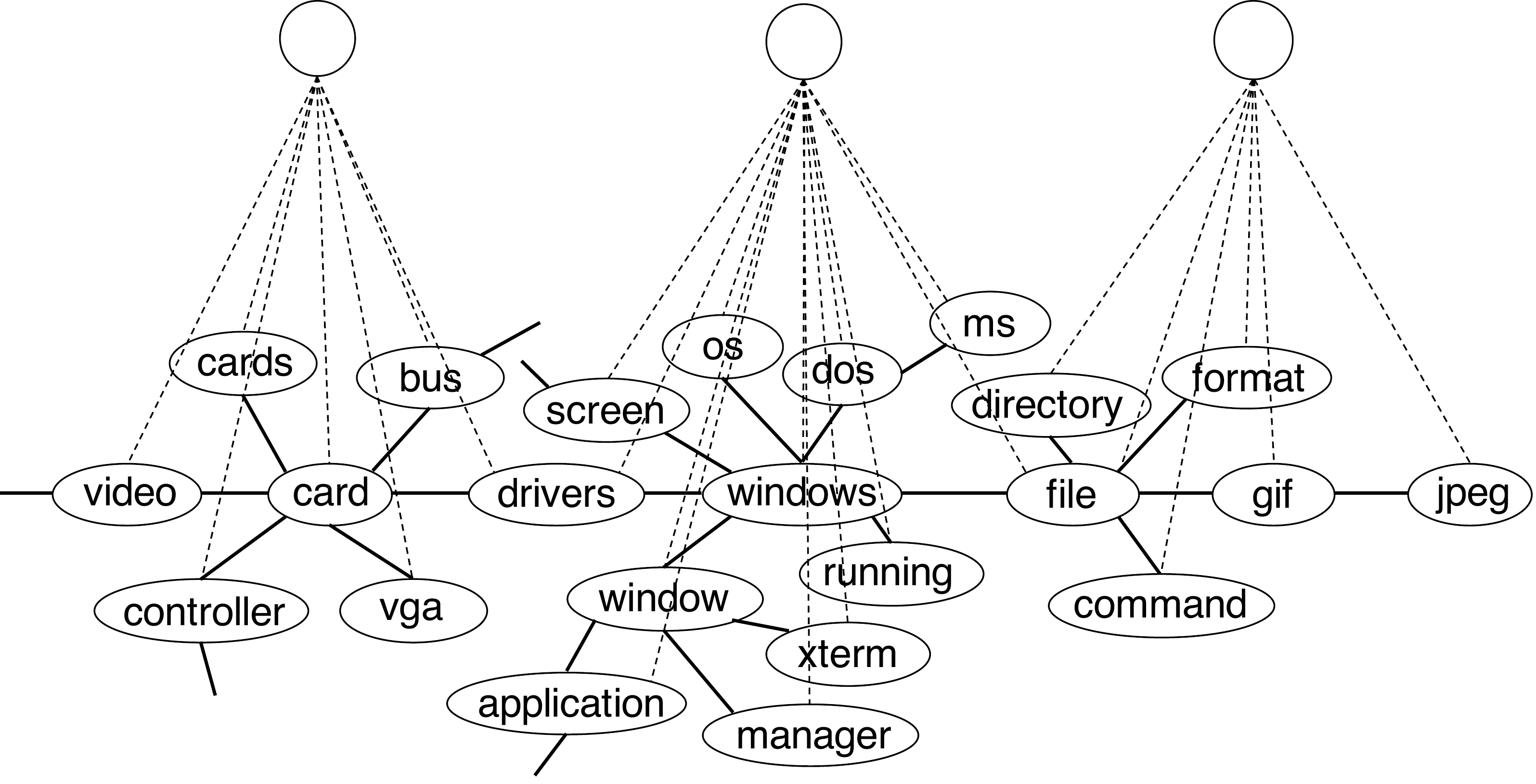}
\end{center}
\caption{Example of local structure in 20newsgroup dataset. Three local neighborhoods are shown: video card, os, file format.}
\label{fig.20news}
\end{figure}

In CNNs, sparse layers are constructed by moving a small sliding window cross the spatial extent of the level below. A neuron is introduced for each position of the sliding window and it is connected to all units within the window at that position. The unspoken word for doing so is that the units in a small spatial region tend to be strongly correlated in their activations.  While there is no spatial information in vector-form data, it is still possible to identify subsets of variables that are strongly correlated, and it is hence still possible to learn sparse architectures. An example is shown in Fig. \ref{fig.20news}, where the observed variables are words from the \textit{20newsgroup} dataset. And it can be seen that three subsets of highly correlated words exist among the observed variables. The statistical structure, in place of spatial or sequential structure, is another promising way to achieve sparse connectivity.

We propose the following method for learn sparse FNNs: (1) Run the Chow-Liu algorithm \cite{chow1968approximating} to learn a tree-structured probabilistic graphical models over the observed features; (2) use the tree to identify subsets of correlated features; introduce a new neuron for each subset; convert the newly introduced neurons into observed variables; and repeat (1) and (2) to build multiple hidden layers. To capture global patterns, we allow a small number of neurons to be connected to all units at the level below. We call deep models built by our method Tree-Receptive-Field networks (\emph{TRF-nets}).
Here are the contributions that we make in this papers:
\begin{itemize}
\item We propose a novel algorithm for learning sparse FNNs.  
\item We have conducted extensive experiments to compare TRF-nets with dense FNNs.  The results show that the TRF-nets achieve better or comparable classification performance with much fewer parameters and sparser structures, and they are more interpretable.
\item We have also empirically compared TRF-nets with sparse FNNs obtained by pruning dense FNNs and models obtained by regularization-based methods. Overall, the TRF-nets outperform other models and have higher interpretability.
\end{itemize}

\section{Related Works}
The primary goal in structure learning is to find a model with optimal or close to optimal generalization performance. Bruce-force search is not feasible because the search space is large and evaluating each model is costly as it necessitates training.  Early works in the 1980's and 1990's have focused on what we call the \textit{micro expansion} approach where one starts with a small network and gradually add new neurons to the network until a stopping criterion is met \cite{kwok1997constructive,kwok1997objective,bello1992enhanced,ash1989dynamic,fahlman1990cascade}. The word ``micro" is used here because at each step only one (or a few) neurons are added. This makes learning a large model computationally difficult as reaching a large model would require many steps and model evaluation is needed at each step. In addition, those early methods typically do not produce layered structures that are commonly used today. Note that, in another line of work, micro expansion methods have been developed for sum-product networks \cite{hsu2017online}.

Recent efforts have concentrated on what we call the \textit{contraction approach} where one starts with a larger-than-necessary structure and reduce it to the desired size. Contraction can be done either by repeatedly pruning neurons and/or connections \cite{Cun90optimalbrain,Hassibi93secondorder,han2015learning,guo2016dynamic,Srinivas2015,li2016pruning}, or by using regularization to force some of the weights to zero \cite{collins2014memory,wen2016learning}. From the perspective of structure learning, the contraction approach is not ideal because it requires a complex model as input. A key motivation for a user to consider structure learning is to avoid building models manually.

A third approach is to explore the model space stochastically. One way is to place a prior over the space of all possible structures and carry out MCMC sampling to obtain a collection of models with high posterior probabilities \cite{adams2010learning}. Another way is to encode a model structure as a sequence of numbers, use a reinforcement meta model to explore the space of such sequences, learn a good meta policy from the sequences explored, and use the policy to generate model structures \cite{baker2016designing,zoph2016neural}.  An obvious drawback of such \textit{stochastic exploration} method is that they are computationally very expensive.

What we propose in this paper is a \textit{macro expansion} method where we start from scratch and repeatedly add layers of hidden units until a threshold is met. Our method is computationally cheaper than the micro expansion approach because it evaluates one two-layer model for each layer of the final model. It also produces layered models. Note that, while parameters are trained layer by layer in deep belief network and deep Boltzmann machines, we learn model structures layer by layer. The learned model is trained as a whole via backpropagation.

Another macro expansion method has recently been proposed by Liu et \cite{liu2017structure}. It learns a new layer by solving a difficult multi-objective optimization problem. In contrast, we run Chow-Liu's algorithm to build a tree among the units on the current layer, which gives a ``spatial structure" among the units, and we build the next layer by applying the concept of receptive field from CNNs on the tree.

\section{Method}
In this section, we present our method for learning sparse FNNs.

\subsection{Learning the Tree Structures}
\begin{figure}[t]
\begin{center}
\includegraphics[width=\linewidth]{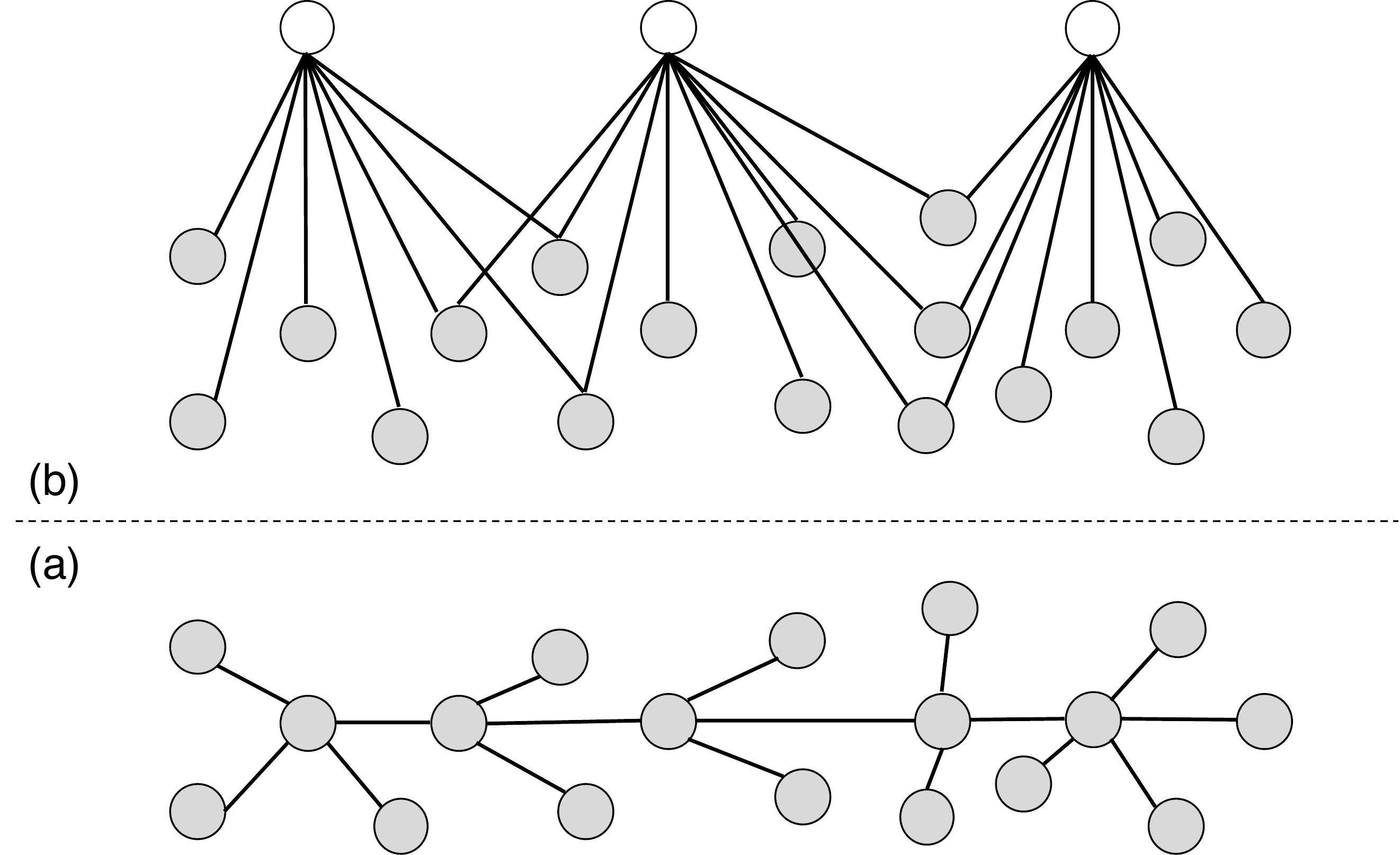}
\end{center}
\caption{Model construction by learning the local structure of observation variables via Chow-liu Tree.}
\label{fig.chowliu}
\end{figure}

We start by learning a tree-structured probabilistic graphical model for the observed variables. Fig. \ref{fig.chowliu}(a) illustrates one such model. Each node in the model represents an observed variable. We assume the observed variables are discrete. In the case of real-valued data, we would discretize the data for the purpose of this subsection and use the original data else where. 

There are multiple ways to parameterize the model. In this paper, we turn the tree into a rooted tree by arbitrarily picking one of the nodes as the root. The parameters of the model then includes the conditional probability $p(x_t|pa(x_t))$ of each variable $x_t$ given its parent $pa(x_t)$. In other words, we deal with a tree-structured Bayesian network here.  The model defines a joint distribution over all the observed variables:
\begin{equation}
p(\mathbf{x}|\mathcal{T}, \theta) = \prod_{t\in \mathcal{V}} p(x_t | pa(x_t)),
\end{equation}
where $\mathcal{T}$ denote the tree structure and $\theta$ denotes the model parameters.

Given a dataset $\mathcal{D}$, the log-likelihood of the parameters $\theta$ and the model structure $\mathcal{T}$ is
\begin{equation}
\begin{aligned}
l(\theta, \mathcal{T}|\mathcal{D}) &= \log p(\mathcal{D}|\theta, \mathcal{T}) \\
&= \sum_t \sum_k N_{tk}\log p(x_t=k|\theta) \\
&+ \sum_{s,t}\sum_{j,k}N_{stjk}\log \frac{p(x_s=j, x_t=k|\theta)}{p(x_s=j|\theta)p(x_t=k|\theta)},
\end{aligned}
\end{equation}
where $N_{stjk}$ is the number of times node $s$ is in state $j$ and node $t$ is in state $k$, and $N_{tk}$ is the number of times node $t$ is in state $k$. The maximized log-likelihood of the model structure $\mathcal{T}$ is
\begin{equation}
\begin{aligned}
l^*(\mathcal{T}|\mathcal{D}) &= \max_{\theta}l(\theta, \mathcal{T}|\mathcal{D}) \\
&= \sum_{t\in \mathcal{V}} \sum_k \hat{p}(x_t=k)\log \hat{p}(x_t=k)\\
&+ \sum_{(s,t)\in \mathcal{E}(\mathcal{T})}\hat{I}(x_s, x_t | \mathcal{D}),
\end{aligned}
\end{equation}
where $\hat{p}(x_t=k)=N_{tk}/N$ is the emperical distribution and $\hat{I}(x_s, x_t | \mathcal{D})$ is the emperical mutual information between $x_s$ and $x_t$, which is given by the following formula:
\begin{equation}
\begin{aligned}
\resizebox{\linewidth}{!}{$\hat{I}(x_s, x_t|\mathcal{D}) = \sum_j \sum_k \hat{p}(x_s=j, x_t=k)\log \frac{\hat{p}(x_s=j, x_t=k)}{\hat{p}(x_s=j)\hat{p}(x_t=k))}.$}
\end{aligned}
\end{equation}
Thus the tree topology that maximize the likelihood can be found by computing the maximum weight spanning tree, where the edge weights are the emperical pairwise mutual information. This is also known as the Chow-Liu algorithm \cite{chow1968approximating}. We use \textit{ChowLiu}($\mathcal{D}$) to denote a subroutine that discretizes data and builds a Chow-Liu tree.

\subsection{Building Two-Layer Structures}
After obtaining a tree structure using Chow-Liu's algorithm, we next build a two-layer structure such as the one shown in Fig. \ref{fig.chowliu}(b).  The tree captures correlation strengths among variables. As a matter of fact, a variable is more strongly correlated with its neighbors than with other variables.  We therefore use local neighborhoods in the tree as receptive fields, and introduce a latent node for each of them. A latent node is connected to all the nodes its receptive field. After latent node introduction, edges between observed nodes are removed.

A receptive field over a tree is defined by a field center $x$, which is one of the nodes, and a field radius $r$. It includes $x$ and all nodes reachable from $x$ in $r$ hops. To cover a tree with receptive fields, we need another parameter, stride $s$, which is the number of hops between the centers of two neighboring receptive fields.  Given $r$ and $s$, we create the first receptive field by randomly picking a node as its center.  At each step after that, we pick the center for the next receptive field by identifying a node that is $s$ hops away one of the existing centers. The process terminates when such a node no long exists.

\begin{figure}[t]
\begin{center}
\includegraphics[width=0.8\linewidth]{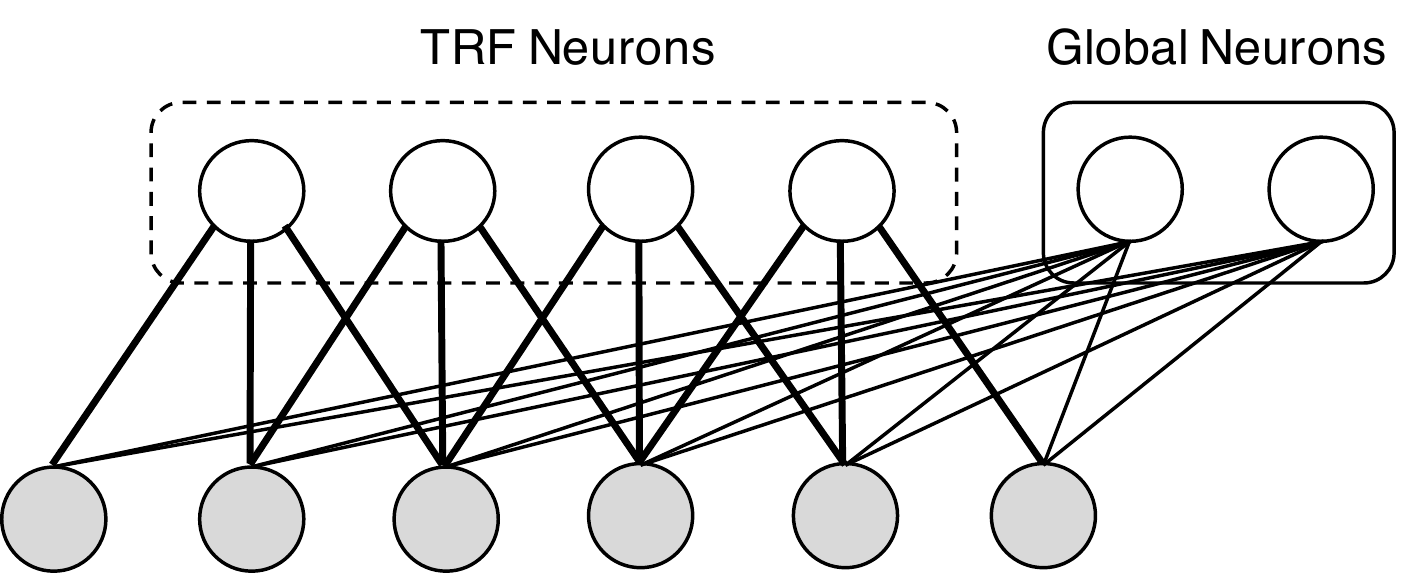}
\end{center}
\caption{A two-layer TRF-net with TRF and global neurons}
\label{fig.localglobal}
\end{figure}

Latent nodes introduced by the method described so far are meant to model local interactions among variables. There could be long-range interactions that they do not capture. Such interactions are indirectly modelled by multiple layers latent nodes. In this paper, we also consider modeling long-range interactions directly. Specifically, we introduce a small number of global neurons and have them connected to all observed nodes. As such, a two-layer structure built by our method looks like what is shown in Fig. \ref{fig.localglobal}. We use \textit{Build2LayerStructure} to denote the procedure described in this subsection. 

\subsection{Training Two-Layer TRF-nets}
After obtaining a two-layer structure, we next turn it into a neural network model and train it as denoising autoencoder.

Let $\mathbf{x}=[x_i]$ and $\mathbf{h}=[h_j]$ be the vectors of visible and hidden units respectively. And let $\mathbf{A}=[a_{ij}]$ be the connectivity matrix where $a_{ij}=1$ if unit $x_i$ is connected to unit $h_j$ and 0 otherwise. Let $\mathbf{W}$ and $\mathbf{b}$ be the weight and bias parameters of the neural network. 

The model defines a conditional distribution $p(\mathbf{h}|\mathbf{x})$ of the hidden units given the observed units:
\begin{equation}
\begin{aligned}
p(\mathbf{h}=\mathbf{1}|\mathbf{x}) = s((\mathbf{A}\circ \mathbf{W})\mathbf{x} + \mathbf{b}),
\end{aligned}
\end{equation}
where $s$ is the probability function, e.g. sigmoid, and $\circ$ denotes element-wise product.
The model also defines a conditional distribution $p(\mathbf{x}|\mathbf{h})$ of the observed units given the hidden units:
\begin{equation}
\begin{aligned}
p(\mathbf{x}|\mathbf{h}) &= Ber(\sigma((\mathbf{A}\circ \mathbf{W})'\mathbf{h}+\mathbf{b})) \quad \text{if $\mathbf{x}$ is binary}\\
p(\mathbf{x}|\mathbf{h}) &= \mathcal{N}((\mathbf{A}\circ \mathbf{W})'\mathbf{h}+\mathbf{b}, \lambda \mathbf{I}) \quad \text{if $\mathbf{x}$ is real-value},
\end{aligned}
\end{equation}
where $(\mathbf{A}\circ \mathbf{W})'$ is the transpose of $\mathbf{A}\circ \mathbf{W}$. 

The model weights are determined using gradient descent and the denoising criteria \cite{vincent2008extracting} is used as the objective function. That is, we first introduce a corruption process $C(\tilde{\mathbf{x}}|\mathbf{x})$ which represents a conditional distribution over corrupted samples $\tilde{\mathbf{x}}$, given a data sample $\mathbf{x}$. The neural network takes a corrupted sample $\tilde{\mathbf{x}}$, maps it to a hidden representation $\mathbf{h} = f(\tilde{\mathbf{x}})=s((\mathbf{A}\circ \mathbf{W})\tilde{\mathbf{x}}+b)$, and then tries to reconstruct the original $\mathbf{x}$, i.e. $\mathbf{x}\rightarrow \tilde{\mathbf{x}} \rightarrow \mathbf{h} \rightarrow \mathbf{x}$. The objective of the denoising training process is
\begin{equation}
\begin{aligned}
\mathcal{L}_{\text{denoising}} = -\mathbb{E}_{\mathbf{x}\sim p_{\text{data}}(\mathbf{x})} \mathbb{E}_{\tilde{\mathbf{x}}\sim C(\tilde{\mathbf{x}}|\mathbf{x})} \log p(\mathbf{x}|\mathbf{h}=f(\tilde{\mathbf{x}})).
\end{aligned}
\end{equation}
With Gaussian distribution, the loss corresponds to reconstruction loss with squared error. With Bernoulli distribution, the loss corresponds to  cross-entropy loss. We use \textit{TrainDenoising} to denote the subroutine for training the two-layer network.

\subsection{Learning a Deep Structure}\label{sec.LearnDeep}
After training the two-layer model, we turn the hidden units into observed variables, and repeat the process to learn multiple layers of hidden units.More specifically, with a learned two-layer model, the observed variables can be mapped to  hidden representations through inference: $p(\mathbf{h}|\mathbf{x})$. Another layer of hidden units can be built on top of the hidden representations. When learning the Chow-Liu tree structure, the hidden representations can be discretized to binary according to whether it activates or does not activate. When learning the weights, the probability values are taken directly as the hidden representations. The resulting layer is then stacked on top of the previous layers. And the procedure is performed recursively.

The overall algorithm of learning TRF-net is shown in Algorithm \ref{algo.localglobal}. The resulting sparse FNN is then trained using backpropagation for upstream tasks.

\begin{algorithm}[t]
	\caption{\textsc{TRF-Net}($\mathcal{D}$, $r$, $s$, $d$)}
	\label{algo.localglobal}
	\DontPrintSemicolon
	\SetAlgoLined

    \KwIn{$\mathcal{D}$---a collection of data, $k$---kernel size, $s$---stride, $d$---depth}
    \KwOut{$m$---TRF-net}
    \BlankLine
    $\mathcal{D}_1 \leftarrow \mathcal{D}$, $m \leftarrow null$\\
    \Repeat{the number of layers in $m$ reaches $d$}{
    	$\mathcal{T} \leftarrow$ \textsc{ChowLiu}($\mathcal{D}_1$) \\
    	$\mathcal{S} \leftarrow$ \textsc{Build2LayerStructure}($\mathcal{T}$, $r$, $s$) \\
    	$m_1 \leftarrow$ \textsc{TrainDenoising}($\mathcal{S}$, $\mathcal{D}_1$) \\
    	$\mathcal{D}_1 \leftarrow$ \textsc{ProjecData}($m_1$, $\mathcal{D}_1$)\\
    	\tcc{End current layer}
    	$m \leftarrow$ \textsc{StackModels}($m$, $m_1$)
    }
    \Return $m$
\end{algorithm}

\begin{table*}
\begin{center}
\begin{small}
\begin{tabular}{l|lrr|c|lr|c}
				& \multicolumn{3}{c|}{TRF-net}   & FNN & \multicolumn{2}{c|}{Pruned FNN} & Reg FNN	 \\  \hline
				&		& \multicolumn{2}{c|}{Parameter \# /}	& &	& & \\
Task			& Accuracy 	& \multicolumn{2}{c|}{Sparsity}	& Accuracy & Accuracy & Sparsity & Accuracy\\ \hline
Tox21 Average		& \textbf{0.8135$\pm$ 0.0038}		& 0.22M / &  10.33\% & 0.8010$\pm$0.0017		& 0.7998$\pm$0.0034		& 32\% & 0.8038$\pm$0.0015\\
AG's News 			& \textbf{91.80\%$\pm$0.05\%}  & 1.62M / &  6.93\%	& 91.61\%$\pm$0.01\%  &  91.49\%$\pm$0.09\%  & 6\% & 91.54\%$\pm$0.02\% \\
DBPedia				& \textbf{98.02\%$\pm$0.01\%}  & 3.50M	/ & 13.90\%	& 97.99\%$\pm$0.04\%	&  97.95\%$\pm$0.02\%	& 25\% & 97.91\%$\pm$0.03\%\\
Yelp Review Full 	& \textbf{59.14\%$\pm$0.06\%}	& 3.23M / & 13.85\%	& 59.13\%$\pm$0.14\%	&  58.83\%$\pm$0.01\%	& 24\% & 59.00\%$\pm$0.11\%\\
Yahoo!Answer		& 71.40\%$\pm$0.01\%	& 2.27M	/ & 10.61\%	& \textbf{71.84\%$\pm$0.07\%}  &  71.74\%$\pm$0.05\%  & 32\% & 71.15\%$\pm$0.05\%\\
Sogou News			& \textbf{96.29\%$\pm$0.07\%}	& 2.27M / & 10.10\%	& 96.11\%$\pm$0.06\%  &  96.20\%$\pm$0.06\%	& 31\% & 95.56\%$\pm$0.04\%\\
\end{tabular}
\end{small}
\end{center}
\caption{Comparison between TRF-nets, FNNs, Pruned FNNs and Reg FNNs on 6 classification datasets. The structures of FNNs are chosen by using validation data. Each experiment is run for three times.}
\label{table.text}
\end{table*}

\section{Experiments}
We have carried out experiments to determine whether our method can produce high quality model structures. The TRF-nets produced by our method are compared with FNNs obtained by manual grid search, and sparse FNNs obtained by weight pruning and weight regularization \cite{han2015learning,collins2014memory}. The micro expansion and stochastic exploration methods are not included in the comparisons because they do not produce layered FNNs and they are extremely expensive computationally. The macro expansion method by [Liu et al 2017] would be a good baseline. Unfortunately, we were able to obtain its implementation.

\subsection{Datasets}
Six publically available datasets were use in the experiments. One  of them is about chemical compounds, and the other five are text data.
\begin{itemize}
\item \textbf{Tox21 challenge dataset.}\footnote{https://github.com/bioinf-jku/SNNs} There are about 12,000 environmental chemical compounds in the dataset. The tasks are to predict 12 different toxic effects. We treat them as 12 binary classification tasks. We filter out sparse features which are present in fewer than 5\% of the compounds, and rescale the remaining 1,644 features to zero mean and unit variance.
We report the average AUC as the results for this dataset.
\item \textbf{Text classification datasets.}\footnote{https://github.com/zhangxiangxiao/Crepe} The five text datasets from~\cite{zhang2015character}: \textit{AG's News}, \textit{DBPedia}, \textit{Yelp Review}, \textit{Yahoo!Answer} and \textit{Sogou News}. The number of classes ranges from 2 to 14 and the dataset sizes are around 130,000$\sim$1,400,000. Stop words were removed during preprocessing. The top 10,000 most frequent words are selected as the vocabulary, and each document is represented as a bag-of-words.
\end{itemize}  

\subsection{Experiment Setup}
We compare TRF-nets with standard fully-connected feed-forward neural networks (FNNs), sparse neural networks obtained by weight pruning (Pruned FNNs) \cite{han2015learning}, and FNNs with L1 weight regularization (Reg FNNs) \cite{collins2014memory}.

When running the TRF-net algorithm on the Tox21 dataset, we used receptive fields with radius $r=5$. Two different values, 1 and 2, were used for the stride $s$.  When $s=1$, the number of units on the next layer is the same as the number of units on the current layer. When $s=2$, the number of units on the next layer is reduced by half. For text data, there are 10,000 input units. So, we set $r=6$ and $s=5$ for the first layer, $r=5$ and $s=4$ for the second layer to quickly reduce the number of units. The number of global neurons introduced at each layer is 10\% of the number of TRF neurons on the same layer.

In the case of FNNs, the number of layers and the number of hidden units were determined by manual grid search. We followed \cite{klambauer2017self} and chose the number of neurons for each layer from \{512, 1024, 2048\}, and tested both rectangular and conic structures. We considered up to 4 hidden layers. For Pruned FNNs, we took the best FNN as the initial model and performed weight pruning as described in \cite{han2015learning}. The pruned model is then retrained to obtain the final model. For Reg FNNs, we added L1 norm term to the training loss with the regularization strength found through validation (around 1e-5). 

In all cases, we used ReLUs \cite{nair2010rectified} as the non-linear activation functions, and Adam \cite{kingma2014adam} as the optimizer, and we applied dropout \cite{srivastava2014dropout} with rate 0.5. We ran all the experiments for three times and report the average results.

\begin{table}
\begin{center}
\begin{tabular}{l|l|lr}
					& TRF-net   & \multicolumn{2}{c}{TRF only}	 \\  \hline
					&				&  & Sparsity \\
Task				&  	Accuracy & Accuracy  	& w.r.t FNNs\\ \hline
Tox21 Average		& 0.8135		& \textbf{0.8196}		& 2.26\% \\
AG's News 			& \textbf{91.80\%}  	& 91.00\%		& 2.48\% \\
DBPedia				& \textbf{98.02\%}  	& 97.75\% 		& 9.94\% \\
Yelp Review Full 	& \textbf{59.14\%}		& 58.69\%		& 9.49\% \\
Yahoo!Answer		& \textbf{71.40\%}		& 70.43\%		& 5.69\% \\
Sogou News			& \textbf{96.29\%}		& 96.03\% 		& 5.42\% \\
\end{tabular}
\end{center}
\caption{Comparison between TRF-nets with global neurons and TRF-nets with TRF neurons only}
\label{table.sparse_path}
\end{table}
\begin{table*}[t]
\begin{center}
\begin{small}
\begin{tabular}{llllllll}
Layers & Tox21 Average & AG's News & DBPedia & Yelp Review Full & Yahoo!Answer & Sogou News \\ \hline
1 & \textbf{0.8139$\pm$0.0026} & 91.65\%$\pm$0.14\% & \textbf{98.09\%$\pm$0.01\%} & 59.00\%$\pm$0.12\% & 71.43\%$\pm$0.02\% & 96.24\%$\pm$0.04\% \\
2 & 0.7992$\pm$0.0020 & 91.82\%$\pm$0.22\% & 98.02\%$\pm$0.01\% & 58.89\%$\pm$0.09\% & 71.42\%$\pm$0.03\% & 96.18\%$\pm$0.02\% \\ 
3 & 0.8135$\pm$0.0038 & 91.80\%$\pm$0.05\% & 98.04\%$\pm$0.02\% & \textbf{59.14\%$\pm$0.06\%} & 71.40\%$\pm$0.01\% & \textbf{96.29\%$\pm$0.07\%}\\
4 & 0.8025$\pm$0.0091 & 91.88\%$\pm$0.14\% & 98.04\%$\pm$0.01\% & 59.09\%$\pm$0.04\% & 71.39\%$\pm$0.07\% & 96.17\%$\pm$0.04\%\\
5 & 0.8081$\pm$0.0063 & \textbf{92.02\%$\pm$0.06\%} & 98.04\%$\pm$0.04\% & 59.08\%$\pm$0.11\% & \textbf{71.51\%$\pm$0.04\%} & 96.20\%$\pm$0.01\%\\
6 & 0.7991$\pm$0.0048 & 92.00\%$\pm$0.14\% & 98.02\%$\pm$0.03\% & 59.01\%$\pm$0.04\% & 71.48\%$\pm$p.02\% & 96.17\%$\pm$0.02\%\\
7 & 0.8055$\pm$0.0101 & 91.88\%$\pm$0.04\% & 97.99\%$\pm$0.03\% & 58.98\%$\pm$0.05\% & 71.38\%$\pm$0.03\% & 96.21\%$\pm$0.03\%
\end{tabular}
\end{small}
\end{center}
\caption{Precision of TRF-nets with different depths on classification datasets. 
For each task, best result is \textbf{bolded}. }
\label{table.depths}
\end{table*}
\subsection{Results}
\subsubsection{Classification Performances}
Table~\ref{table.text} shows the classification performances of TRF-nets, FNNs, Pruned FNNs and Reg FNNs on different datasets. It is clear that TRF-nets achieved better AUC scores than FNNs on the Tox21 dataset and better or comparable accuracy on 5 text classification datasets. They did so with significantly fewer parameters and sparser structures. In comparison with FNNs, the sparsity of TRF-nets (\# of connections w.r.t that of fully-connected ones) is around 10\% for all datasets. This confirms that our method is able to achieve high generalization performance with much fewer parameters and sparser structures.
Compared with Pruned FNNs and Reg FNNs, TRF-nets also achieved better or comparable classification accuracy. Prunned FNNs have comparable amounts of model parameters as TRF-ntes. Reg FNNs are not sparse models. Their weights are pushed toward 0 by regularization, but few actually reached 0. With weights of absolute values less than 0.001 removed, their sparsity is between 20\% and 30\% in comparison with FNNs.  Note that pruned FNNs and Reg FNNs were obtained from FNNs that took much manual search to construct. In contrast, TRF-nets were automatically and directly learned from data.

\subsubsection{Contribution of the Tree Receptive Fields}
To validate our assumption that the TRF neurons in TRF-nets capture most of the patterns in data, we performed another set of experiments with all the global neurons removed. Table~\ref{table.sparse_path} shows the results. As we can see, the classification performances degraded only slightly, while model sparsity is futher improved significantly. The results not only show the importance of the TRF neurons in TRF-nets, but also shows that our structure learning method is effective.

\subsubsection{Influence of the Model Depth}
The results reported in Tables \ref{table.text} and \ref{table.sparse_path} were obtained using TRF-nets with 3 hidden layers. The decision to use 3 hidden layers was influenced by the literature. Typically authors use only a small number of hidden layers when it comes to FNNs. Nonetheless, it would be interesting to see how model depth influences the performance TRF-nets. So, we performed a set of experiments with varying model depths. The results are shown in Table~\ref{table.depths}. It can be seen that our method continues to work when the network become deeper. For example, for the \textit{AG's News} dataset, the best performance was achieved with a 5-layer TRF-net. In contrast, deep architecture might lead to severe overfitting for FNNs.

\subsubsection{Computational Time}
A comparison of the computational time for TRF-nets and Pruned FNNs is given in Table \ref{table.computation}. As it can be seen, the structure learning phase of TRF-nets costs almost as much time as the finetuning phase. The overall computational time is comparable with Pruned FNNs without structure searching. Note that FNNs, Pruned FNNs and Reg FNNs need to determine the best-performing FNN structure by grid searching in the pre-defined structure space, as described in the experiment setup. Pruned FNNs further prune and finetune the best structure. Although for TRF-nets there are also hyperparameters, i.e. receptive field and stride, it is significantly easier to determine those controlling hyperparameters than to determine the actual structures. Therefore, in practice the overall computational time including grid searching over the structure space for FNNs, Pruned FNNs and Reg FNNs is significantly larger than that for TRF-nets. All neural networks are trained on a Tesla K20 GPU, while the structure learning is run on a CPU.

\begin{table}
\begin{center}
\begin{tabular}{l|cc|cc}
				& \multicolumn{2}{c|}{TRF-net}  & \multicolumn{1}{c}{Pruned FNN}\\  \hline
				& & & w/o structure \\
Task			& structure 	& finetune	& searching \\ \hline
Tox21 Average		& 145 & 27 & 43 \\
AG's News 			& 3,388 & 1,670 &  3,956 \\
DBPedia				& 4,082 & 2,829 &  6,562 \\
Yelp Review Full 	& 5,072 & 3,135 &  3,618 \\
Yahoo!Answer		& 7,215 & 5,870 & 12,440 \\
Sogou News			& 4,311 & 3,948 &  7,340 \\
\end{tabular}
\end{center}
\caption{Comparison of computational time (s) between TRF-nets and Pruned FNNs on 6 classification datasets.}
\label{table.computation}
\end{table}

\subsubsection{Interpretability}
Next we compare TRF-nets with FNNs and Pruned FNNs in terms of the interpretability of their hidden unit. We characterize the ``meaning" of a hidden unit using the top 10 words that have the strongest correlations with the unit. Following \cite{chen2017sparse}, we measure the ``interpretability" of a hidden unit by considering how similar pairs of words in the top-10 list are using a word2vec model \cite{mikolov2013efficient,DBLP:conf/nips/MikolovSCCD13} trained on part of the Google News datasets. The interpretability score of a model is defined as the average of interpretability scores of all top-layer units. 
The results are reported in Table \ref{table.interpretscore}. There are no results for the \textit{Sogounews} dataset because its vocabulary are Chinese pingyin characters and most of them do not appear in the Google News word2vec model. For the fairness of comparison, all models have approximately the same number of top-layer hidden units. As it can be seen, TRFT-nets outperformed FNNs in most cases. In contrast, Pruned FNNs achieved lower scores than FNNs in most cases. This indicates that the retraining after weight pruning reduces the interpretability of hidden units.
As concrete examples, Table \ref{table.interpret_qualitative} shows the top 3 words that were used to characterize some of the hidden units. Those word groups are clearly meaningful. For example, the first group for the \textit{Yahoo!Answer} dataset are about computers, while the third group is about medicine. 

\begin{table}[t]
\begin{center}
\begin{tabular}{llll}
Task & TRF-nets & FNNs & Pruned FNNs \\
\hline
Yelp Review Full 	&  \textbf{0.1176} &  0.1117 & 0.1002\\
DBPedia				&  \textbf{0.0577} &  0.0497 & 0.0553\\
Yahoo!Answer		&  \textbf{0.1754} &  0.1632 & 0.1553\\
AG's News 			&  0.0527 & \textbf{0.0595} & 0.0561\\
\end{tabular}
\end{center}
\caption{Interpretability scores of TRF-nets, FNNs and Pruned FNNs on different datasets}
\label{table.interpretscore}
\end{table}
\begin{table}
\begin{center}
\begin{tabular}{ll}
Task & Characterization Words \\
\hline
\multirow{3}{*}{Yelp Review Full} & tasteless unseasoned flavorless\\ 
& paprika panko crusts unagi  crumb \\ 
& vindaloo tortas spicey wink drapes \\
\hline
\multirow{3}{*}{DBPedia} & album songwriting chet saxophone \\ 
& hurling backstroke badminton skier \\ 
& journalists hardcover editors \\
\hline
\multirow{3}{*}{Yahoo!Answer} & xp desktop adobe peut cpu \\ 
& pagans atheist mormons apostle \\ 
& tylenol diabetic acids kidneys \\
\hline
\multirow{3}{*}{AG's News} & mozilla mainframe  microprocessors \\ 
&  republicans prosecutor argument jfk \\ 
& noted furious harsh concessions \\
\hline
\end{tabular}
\end{center}
\caption{Qualitative interpretability results of hidden units in TRF-nets. Each line corresponds to one hidden unit.}
\label{table.interpret_qualitative}
\end{table}

\section{Conclusions}
Structure learning for deep neural network is a challenging and interesting research problem. We have proposed an unsupervised structure learning method, which first learns a tree-structured probabilistic graphical model and then constructs hidden neurons with local receptive field over the tree. The resulting TRF-nets have shown to achieve better or comparable classification performance in all kinds of tasks compared with standard FNNs, while containing significantly fewer parameters.
In addition, TRF-nets have also shown to be more interpretable than FNNs, which is interesting because interpretability is an important issue in deep learning. 

\section*{Acknowledgments}
Research on this article was supported by Hong Kong Research Grants Council under grant 16212516.

\bibliographystyle{named}
\bibliography{reference}

\end{document}